\documentclass[manuscript]{acmart} %

\usepackage{booktabs} 
\usepackage{algpseudocode,algorithm,algorithmicx}
\usepackage{hyperref}

\DeclareMathOperator*{\argmin}{arg\,min}

\usepackage[acronym,toc,shortcuts]{glossaries}
\makenoidxglossaries                    

\usepackage{subfigure}
\usepackage{flafter}
\usepackage{float}

\newacronym{MCA-ann}{CoMaC-ann}{CoMaC annealing}
\newacronym{MCA-seq}{CoMaC-seq}{CoMaC sequential}

\newacronym{NMI}{NMI}{Normalized Mutual Information}

\newacronym{CEC}{CEC}{Cross-Entropy Clustering}
\newacronym{IB}{IB}{Information Bottleneck}
\newacronym{CEC-IB}{CEC-IB}{cross-entropy clustering with information bottleneck regularization}

\newacronym{MCA}{CoMaC}{Constrained Markov Clustering}

\newacronym{GMM}{GMM}{Gaussian Mixture Model}

\newacronym{iid}{i.i.d}{independent and identically distributed}

\newacronym[shortplural=RVs, longplural=Random variables]{RV}{RV}{Random variable}
\newacronym[shortplural=PMFs, longplural=probability mass functions]{PMF}{PMF}{probability mass function}

\newacronym{KLDR}{KLDR}{Kullback-Leibler divergence rate}

\newacronym{DFS}{DFS}{depth-first search}                  

\setcopyright{acmcopyright}

\acmDOI{xx.xxx/xxx_x}

\acmISBN{978-1-4503-8713-2/22/04}

\acmConference[SAC'22]{ACM SAC Conference}{April 25 –April 29, 2022}{Brno, Czech Republic}
\acmYear{2022}
\copyrightyear{2022}

\acmArticle{4}
\acmPrice{15.00}

\setcopyright{none}

\begin{document}
\title{Semi-Supervised Clustering via Information-Theoretic Markov Chain Aggregation}
  
\renewcommand{\shorttitle}{Semi-Supervised Clustering via Markov Chain Aggregation}

\author{Sophie Steger}
\orcid{1234-5678-9012}
\affiliation{%
  \institution{Signal Processing and Speech Communication Laboratory, Graz University of Technology}
  \streetaddress{Inffeldgasse 16c}
  \city{Graz} 
  \country{Austria} 
  \postcode{8010}
}
\email{sophie.steger@student.tugraz.at}

\author{Bernhard C. Geiger}
\affiliation{%
  \institution{Know-Center GmbH}
  \streetaddress{Inffeldgasse 13}
  \city{Graz} 
  \country{Austria} 
  \postcode{8010}
}
\email{geiger@ieee.org}

\author{Marek \'Smieja}
\affiliation{%
  \institution{Faculty of Mathematics and Computer Science, Jagiellonian University}
  \streetaddress{\L{}ojasiewicza 6}
  \city{Krakow} 
  \country{Poland}
  }
\email{marek.smieja@uj.edu.pl}

\renewcommand{\shortauthors}{S. Steger et al.}

\begin{abstract}
We connect the problem of semi-supervised clustering to constrained Markov aggregation, i.e., the task of partitioning the state space of a Markov chain. We achieve this connection by considering every data point in the dataset as an element of the Markov chain's state space, by defining the transition probabilities between states via similarities between corresponding data points, and by incorporating semi-supervision information as hard constraints in a Hartigan-style algorithm. 
The introduced \gls{MCA} is an extension of a recent information-theoretic framework for (unsupervised) Markov aggregation to the semi-supervised case. Instantiating \gls{MCA} for certain parameter settings further generalizes two previous information-theoretic objectives for unsupervised clustering. Our results indicate that \gls{MCA} is competitive with the state-of-the-art.
\end{abstract}

%
%



\maketitle

\newcommand{\ent}[1]{H(#1)}
\newcommand{\mutinf}[1]{I(#1)}

\newcommand{\numitermax}{\text{\#iter}_{\max}}
\newcommand{\numiter}{\text{\#iter}}
\newcommand{\NMI}[2]{ \text{NMI}(#1, #2) }
\newcommand{\ginit}{g_{\text{init}}}
\newcommand{\Mcal}{\mathcal{M}}
\newcommand{\Ncal}{\mathcal{N}}
\newcommand{\Minit}{\Mcal_{\text{init}}}
\newcommand{\Ninit}{\mathcal{N}_{\text{init}}}
\newcommand{\Xcal}{\mathcal{X}}
\newcommand{\Ycal}{\mathcal{Y}}
\newcommand{\Xvec}{\mathbf{X}}
\newcommand{\cost}{\mathcal{C}}
\newcommand{\Pvec}{\mathbb{P}}
\newcommand{\card}[1]{|#1|}
\newcommand{\betatar}{\beta_{\text{target}}}
\newcommand{\e}[1]{\mathrm{e}^{#1}}

\newcommand{\figc}[4]{\begin{figure}[thbp]\centering\includegraphics[#1]{#2}\caption{#3}\label{fig:#4}\end{figure}}%

\def\our{CoMaC}
\newcommand\change[1]{\color{red}#1}
\newcommand\new[1]{\color{green}#1}

\begin{figure}\centering
\subfigure[]{\includegraphics[width=0.3\textwidth]{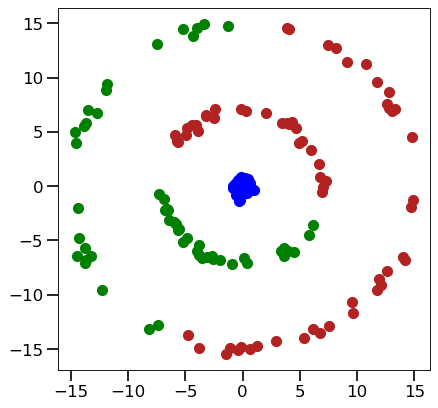}}
\subfigure[]{\includegraphics[width=0.3\textwidth]{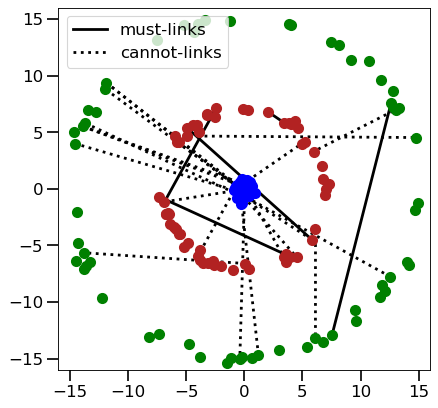}}
\caption{Circles dataset with three concentric circles consisting of 60 data points each. Result for (unsupervised) Markov aggregation clustering (a) and the proposed semi-supervised \gls{MCA} with 30 constraints (b). While also the unsupervised approach can learn non-linear boundaries between clusters, the addition of pairwise constraints helps avoiding bad local optima.}
\label{fig:circles}
\end{figure}

\section{Introduction}

A popular approach to clustering, especially if only pairwise similarities between data points are available, is to view the problem from the perspective of random walks. From this perspective, each data point is represented by a state in the state space of a Markov chain, whose transition probabilities are determined by the pairwise similarities between the corresponding data points. The clustering problem can then be solved via aggregating the state space of the thus defined Markov chain. In this paper, we focus on the unifying framework from~\cite{Amjad_2020}, which captures previous information-theoretic approaches to Markov aggregation~\cite{Meyn_MarkovAggregation,Xu_Reduction,Vidyasagar_MarkovAggregation,GeigerEtAl_OptimalMarkovAggregation} and, if instantiated appropriately, clustering~\cite{ALUSH2016284, tishby_dataclustering} as special cases (see Section~\ref{sec:related}). 

Although clustering via Markov aggregation has a solid theoretical basis, allows for creating non-linear decision boundaries, and was shown to achieve competitive performance~\cite{ALUSH2016284}, it appears to be highly sensitive on a careful selection of hyperparameters or optimization procedures. In random walk-based clustering, even representing a dataset as a Markov chain requires appropriately selecting hyperparameters, cf.~\eqref{eq:transform} below. Tuning these hyperparameters to individual datasets is cumbersome and severely limits the practical applicability of the respective clustering method. Moreover, there are no clear rules and objective evaluation measures for their selection because of the unsupervised nature of clustering.

In this paper, we propose Constrained Markov Clustering (\gls{MCA}), the extension of clustering via Markov aggregation~\cite{Amjad_2020} to the semi-supervised setting, where the side information is given in the form of partition-level information \cite{k_means} (some data points are labeled by their cluster index) or pairwise constraints \cite{wagstaff2001constrained} (for some pairs of data points we know whether they belong to the same or to different clusters, see~\autoref{fig:circles}). 

Experimental results confirm that the proposed adaptions to the Hartigan-style clustering algorithm of~\cite{Amjad_2020} achieve performance on common benchmarks that is competitive with the state-of-the-art in semi-supervised clustering. Furthermore, there are indications that introducing side information makes the algorithm more robust to the selection of some of its hyperparameters. Reducing the sensitivity to hyperparameters is important for (semi-supervised) clustering because the limited available information about class labels typically precludes performing proper validation.

\section{Problem Statement and Related Work}
\label{sec:related}
We address the (semi-supervised) clustering problem via transforming it into a Markov chain aggregation problem. In this section, we thus give a statement of these problems and review the relevant literature to prepare the stage for our method in Section~\ref{sec:method}.

Clustering refers to the task of grouping the elements of a dataset $\Xcal=(x_1,\dots,x_N)$, $x_i\in\mathbb{R}^n$, such that the data points within each group have a higher similarity with each other than with those of different groups, where similarity has to be defined appropriately. If this grouping is deterministic, then there exists a clustering function $g{:}\ \Xcal\to\{1,\dots,K\}$ that maps each element of $\Xcal$ to the index of one of the $K$ clusters. Clustering is successful if the candidate clustering function $g$ is close (in a well-defined sense) to the function $g^\bullet{:}\ \Xcal\to\{1,\dots,K^\bullet\}$ determining the true partition.

Semi-supervised clustering simplifies the task by providing additional information in one of two flavors: First, partition-level side information refers to a subset $\Xcal'$ of $\Xcal$ for which the true cluster indices are known, i.e., $\{(x,g^\bullet(x)) \mid  x\in\Xcal'\subset\Xcal\}$. Such partition-level side information was proposed for k-means~\cite{k_means}, fuzzy c-means~\cite{fuzzy_1,fuzzy_2}, \glspl{GMM}~\cite{r_mixmod}, or \gls{CEC-IB}~\cite{smieja_2017}. The second option are pairwise constraints, which indicate which pairs of data points of $\Xcal$ must or must not be put in the same cluster; this setting is often referred to as constrained clustering. Pairwise constraints are given as
\begin{subequations}\label{eq:pairwise_constraints}
\begin{align}
    \Mcal = \{ \, (x,x') \mid g^\bullet(x) = g^\bullet(x')  \} \\
    \Ncal = \{ \, (x,x') \mid g^\bullet(x) \neq g^\bullet(x')  \}
\end{align}
\end{subequations}
for a (small) subset of pairs $(\Mcal\cup\Ncal)\subset\Xcal^2$. Pairwise constraints have been utilized for discriminative clustering with graph regularization~\cite{SMIEJA201824}, \glspl{GMM}~\cite{cGMM}, or spectral clustering~\cite{spec}.

A special instance of clustering is the problem of Markov aggregation, i.e., the problem of clustering states of a Markov chain. Mathematically, if the stochastic process $X=(X_1,X_2,\dots)$ is an aperiodic, irreducible, and stationary Markov chain (see~\cite{Kemeny_FMC} for terminology) with finite state space $\mathcal{Z}$, then the task is to find an aggregation function $h{:}\ \mathcal{Z}\to\{1,\dots,K\}$ such that the aggregated process $Y=(h(X_1),h(X_2),\dots)$ satisfies certain properties. Several information-theoretic cost functions have been proposed for this problem. For example, the authors of~\cite{Meyn_MarkovAggregation,Vidyasagar_MarkovAggregation} aimed at a maximally predictable process $Y$ by maximizing the mutual information $I(Y_1;Y_2)$, while the authors of~\cite{GeigerEtAl_OptimalMarkovAggregation} selected $h$ such that $Y$ is as Markov as possible as measured via the Kullback-Leibler divergence rate. Recently, an information-theoretic framework for Markov aggregation has been proposed, which aims at finding a minimizer of~\cite{Amjad_2020}
\begin{equation}\label{eq:combined_cost}
  \cost_\beta(X, h)=(1-2\beta)\left(\ent{Y_2|Y_1} - \ent{Y_2|X_1}\right) - \beta \mutinf{Y_1;Y_2}  
\end{equation}
where $H$ denotes the entropy, where the minimum is taken over all functions $h: \mathcal{Z} \to \{1,\dots,K\}$, and where the first two and the third terms represent the operational goals of preserving the Markov property and the temporal dependence structure of $X$, respectively. It can be shown that this framework covers the cost functions of~\cite{GeigerEtAl_OptimalMarkovAggregation},~\cite{Meyn_MarkovAggregation,Vidyasagar_MarkovAggregation}, and~\cite{Xu_Reduction} as special cases for $\beta$ being equal to 0, 0.5, and 1, respectively. 

Markov aggregation is thus presented as a clustering problem. Conversely, by identifying each element of a dataset $\Xcal$ with a state of a Markov chain and by parameterizing the transition probability between states via the similarity of corresponding data points, the clustering problem can be formulated as a Markov aggregation problem. For example, if $d{:}\ \Xcal^2\to [0,\infty)$ is a measure of dissimilarity between data points, then $\Xcal$ can be clustered  via aggregating the Markov chain $X=(X_1,X_2,\dots)$ with state space $\Xcal$ and transition probability matrix $\Pvec=[P_{i,j}]$,
\begin{equation}\label{eq:transform}
    P_{i,j} \propto \e{-\frac{d(x_i,x_j)}{\sigma^2}}
\end{equation}
where $\sigma^2$ is a scaling factor. The candidate aggregation function $h$ obtained by solving~\eqref{eq:combined_cost} can then be interpreted as the candidate clustering function $g$. Indeed, the authors of~\cite{ALUSH2016284} proposed maximizing $\mutinf{Y_1;Y_2}$, where $d(x_i,x_j)$ is 0 if $x_i$ and $x_j$ are $k$-nearest neighbors of each other and $\infty$ otherwise. Furthermore, in~\cite{tishby_dataclustering} $d$ was chosen as the Euclidean distance, $\sigma^2$ as the $k$-nearest neighbor distance, and the authors proposed to minimize
\begin{equation}
    \mutinf{Y_1;X_1} - \beta\mutinf{Y_1;X_{T+1}}
\end{equation}
where $T$ is selected such that the Markov chain $X$ has relaxed to a meta-stable state. These two approaches of~\cite{ALUSH2016284} and~\cite{tishby_dataclustering} (for $T=1$ and symmetric dissimilarity measures $d$)  thus correspond to solving~\eqref{eq:combined_cost} for $\beta=0.5$ and, for, $\beta=1$, respectively.\footnote{For symmetric dissimilarity measures, the transition probability matrix resulting from~\eqref{eq:transform} is reversible, i.e., $\mutinf{Y_1;X_2}=\mutinf{Y_2;X_1}$, cf.~\cite[Sec.~IV.C]{Amjad_2020}.} The main differences between~\cite{tishby_dataclustering,ALUSH2016284} and minimizing~\eqref{eq:combined_cost} rely on the definition of $\Pvec$ in~\eqref{eq:transform} and, potentially, the relaxation time $T$ proposed in~\cite{tishby_dataclustering}.

Preliminary experiments suggest that good clustering performance can only be achieved by the methods in~\cite{tishby_dataclustering,ALUSH2016284} if the parameters $\sigma^2$ (or its proxy $k$) and $T$ are carefully set. Our research hypothesis, which will be confirmed in this paper, is that the provision of pairwise constraints makes the proposed methods less sensitive to these parameter settings.

\section{Semi-Supervised Clustering via Markov Aggregation}
\label{sec:method}
In this section, we introduce our approach to semi-supervised clustering based on Markov aggregation (\gls{MCA}). To this end, we utilize the Markov aggregation problem~\eqref{eq:combined_cost} proposed in~\cite{Amjad_2020} and apply it to a Markov chain $X$ with a transition probability matrix $\Pvec$ depending on the clustering dataset $\mathcal{X}$ via~\eqref{eq:transform}. Specifically, we choose $d$ to be the squared Euclidean distance and $\sigma_k^2$ as the average squared Euclidean distance between the data point and its $k$ nearest neighbors (averaged over all data points), i.e.,
\begin{equation}\label{eq:our_transform}
    P_{i, j} \propto \e{-\frac{\Vert x_i-x_j\Vert^2_2}{\sigma_k} }.
\end{equation}

The authors of~\cite{Amjad_2020} proposed a Hartigan-style algorithm for solving the optimization problem in~\eqref{eq:combined_cost} for a deterministic clustering function $g{:}\ \Xcal\to\{1,\dots,K\}$. Starting from an initial clustering of $\Xcal$ into $K$ clusters, each data point $x$ is mapped to every aggregate state $y \in \{1,\dots,K\}$ and the cost function is evaluated. The data point is then assigned to the aggregate state that minimizes the cost function. Since this algorithm tends to get stuck in poor local optima for small values of $\beta$, an additional annealing procedure was introduced in~\cite{Amjad_2020} that provides clustering functions $g$ obtained for higher values of $\beta$ as initial clusterings for lower values of $\beta$. 

In this work, the algorithm of~\cite{Amjad_2020} is extended in order to accept pairwise constraints $\Mcal$ and $\Ncal$ as given in~\eqref{eq:pairwise_constraints}. Since partition-level side information can easily be converted to pairwise constraints (but not vice-versa), the resulting algorithm can handle both types of side information. Below we describe the initialization, iteration, and annealing procedures of \gls{MCA}.

\paragraph{Initialization} First, the candidate partition function $g$ is initialized such that all pairwise constraints are satisfied. This is done via solving a graph coloring problem, where no adjacent vertices of a graph are allowed to be of the same color. In our procedure, each vertex of this graph either corresponds to an individual data point not involved in any must-link constraint, or to a set of data points that are connected via must-link constraints, while each edge of this graph corresponds to a cannot-link constraint. The initial coloring of the graph is performed by a greedy algorithm (see Algorithm \ref{alg:coloring}) where each vertex is assigned the first color available in sequence. To avoid the algorithm getting stuck in bad local minima, vertices with no cannot-link constraints are assigned a random color. 

\begin{algorithm}
\caption{Greedy coloring algorithm.}
\label{alg:coloring}
\begin{algorithmic}[1]
\Function{$g = $ greedyColoring}{must-link constraints $\Mcal$, cannot-link constraints $\Ncal, K$}
\For{all elements $x \in \Xcal$}
    \State $\Mcal_x = $ \textsc{funcMust}($\Mcal, \Ncal, x$)
    \State $\Ncal_x = $ \textsc{funcCannot}($\Mcal, \Ncal, x$)
    \If{$\Ncal_x$ is empty}
        \State $g(\Mcal_x) \gets $ random value out of $K$ colors
    \Else
        \State $g(\Mcal_x) \gets $ firstColorAvailable($\Ncal_x$)
    \EndIf
\EndFor
\EndFunction
\end{algorithmic}
\end{algorithm}

\paragraph{Iteration} Once the initial partition function is defined, the sequential algorithm minimizes the cost function in~\eqref{eq:combined_cost} iteratively. Cannot-link constraints are incorporated by restricting the possible states of the aggregation function in line~\ref{line:occurrences} of Algorithm \ref{alg:myalgo-seq}. Data points connected by must-link constraints are assigned to an aggregate state simultaneously. Erroneous or noisy pairwise constraints can lead to the case where a data point cannot be assigned to any aggregate state. Then, the aggregate state with the least occurrences of cannot-link constraints is selected. Algorithm~\ref{alg:myalgo-seq} contains the sequential algorithm in~\cite{Amjad_2020} as special case if $\Mcal=\Ncal=\emptyset$ and if the propagation functions are such that $\Mcal_x=\{x\}$ and $\Ncal_x=\emptyset$.

\begin{algorithm}
\caption{Sequential Generalized Information-Theoretic Markov Aggregation with Pairwise Constraints.}
\label{alg:myalgo-seq}
\begin{algorithmic}[1]
\Function{$g = $CoMaC-seq}{$\Pvec$, $\beta$, $K$, $\numitermax$, \textit{optional}: initial aggregation function $\ginit$, must-link constraints $\Mcal$, cannot-link constraints $\Ncal$}
\If{$\ginit$ is empty} \Comment{Initialization}
    \State $g =$ \textsc{greedyColoring}($\Mcal$, $\Ncal$, $K$) 
\Else 
    \State $g \gets \ginit$
\EndIf
\State $\numiter \gets 0$
\While{$\numiter < \numitermax$} \Comment{Main Loop}
    \For{all elements $x \in \Xcal$} \Comment{Optimizing $g$}
        \State $\Mcal_x = $ \textsc{funcMust}($\Mcal,\Ncal, x$) 
        \State $\Ncal_x = $ \textsc{funcCannot}($\Mcal, \Ncal, x$) 
        \State $\Ycal_{pos} = \{1,\dots,K\}\setminus g(\Ncal_x)$ \Comment{Possible aggregate states}
        \If{$\Ycal_{pos}$ is empty}
            \State $\Ycal_{pos} = \argmin\limits_y \card{ y \in g(\Ncal_x) }$ \Comment{Select state with the least occurrences} \label{line:occurrences}
        \EndIf
        \For{all possible aggregate states $y \in \Ycal_{pos}$}
            \State $ g_y(x^\prime) = \begin{cases} g(x^\prime) , & x^\prime \notin \Mcal_x\\
            y , & x^\prime \in \Mcal_x \end{cases} $ 
            \State $C_{g_y} = \cost_\beta(X, g_y)$
        \EndFor
        \State $g = \argmin\limits_{g_y} C_{g_y}$ \Comment{(break ties)}
    \EndFor
    \State $\numiter \gets \numiter+1$
\EndWhile
\EndFunction
\end{algorithmic}
\end{algorithm}

\paragraph{Annealing} The annealing procedure for $\beta$ was introduced in~\cite{Amjad_2020} to avoid the sequential algorithm getting stuck in poor local minima for small values of $\beta$. Annealing is initialized with $\beta=1$, and the resulting aggregation functions are iteratively used as initialization for the sequential algorithm with reduced $\beta$. The annealing procedure itself was not adapted and is shown in Algorithm \ref{alg:myalgo-ann}.

\begin{algorithm}
 	\caption{$\beta$-Annealing Information-Theoretic Markov Aggregation with Pairwise Constraints.}\label{alg:myalgo-ann}
 	\begin{algorithmic}[1]
        \Function {$g =$ CoMaC-ann}{$\Pvec$, $\beta_\text{target}$, $K$, $\numitermax$, $\Delta$, \textit{optional}: must-link constraints $\Mcal$, cannot-link constraints $\Ncal$}
        \State $g$ = \textsc{CoMaC-seq}($\Pvec$, 1, $K$, $\numitermax$, \textit{optional}: $\Mcal$, $\Ncal$) \Comment{Inizialization}
 		\While {$\beta > \betatar$}
 		\State $\beta \gets \max\{\beta - \Delta,\betatar\}$ 
        \State $g$ = \textsc{CoMaC-seq}($\Pvec$, $\beta$, $K$, $\numitermax$, $g$, \textit{optional}: $\Mcal$, $\Ncal$)
 		\EndWhile
 		\EndFunction
 	\end{algorithmic}
 \end{algorithm}

\paragraph{Constraint Propagation} Usually, the sets $\Mcal$ and $\Ncal$ of pairwise constraints are not exhaustive. For example, if $(x,x')\in\Mcal$ and $(x,x'')\in\Mcal$, then also $x'$ and $x''$ must link, even though such relation does not appear explicitly in the side information. If the sets of pairwise constraints are exhaustive, vertices connected by must-link constraints form graph cliques such that every pair of vertices within the clique is connected by a must-link constraint. 
In the case of sparse pairwise constraints, must-link constraints do not necessarily form cliques. To account for this, constraints in $\Mcal$ can be \emph{propagated}. However, propagating these constraints is a non-trivial problem, especially if the pairwise constraints are conflicting (e.g., due to labeling errors). 
In this work we assume non-contradictory constraints. We look for all connected components in the graph given by $\Mcal$ using a \gls{DFS} algorithm. Initially, all vertices of the graph are marked as unvisited. Starting at an arbitrary vertex, we determine all connected vertices and mark them as visited. Iteratively, the procedure is repeated for those vertices until we reach a vertex that has been visited before. Using these connected components, the set $\Mcal$ can be extended, such that it describes a graph consisting of independent cliques only.  
In Algorithm~\ref{alg:coloring}, this is done in the function \textsc{funcMust}. The function returns all $x'$ that must be in the same clusters as $x$ given by the \gls{DFS} algorithm on $\Mcal$.
Additionally, the function \textsc{funcCannot} considers the case where two data points from different cliques are connected by a cannot-link constraint. Then, all elements of the two respective cliques are connected by cannot-link constraints, and thus are not allowed to be in the same cluster. E.g., if $(x,x')\in\Mcal$ and $(x,x'')\in\Ncal$, then also $x'$ and $x''$ should not link, despite $(x',x'')$ not being a labeled cannot-link constraint. Our function \textsc{funcCannot} thus returns $\Ncal_x=\{x',x''\}$.

\section{Experiments}

In this section, we experimentally evaluate the performance of \gls{MCA}\footnote{The data and code used for these experiments is publicly available at \url{https://github.com/stegsoph/Constrained-Markov-Clustering}}. First, we verify the research hypothesis that the introduction of pairwise constraints makes Markov aggregation-based clustering less sensitive to the choice of hyperparameters. Next, we compare its performance with the state-of-the-art semi-supervised clustering techniques following the experimental setup of \cite{smieja_2017}. Finally, we demonstrate and discuss the current limitations of \gls{MCA}.

We measure the accuracy of the obtained clusterings with the \gls{NMI}. It is defined by the mutual information between the true partition $g^\bullet$ and estimated partition $g$ normalized by sum of the entropy of the partitions, i.e., 
\begin{align}
            \NMI{g^\bullet(U)}{g(U)} = \frac{2I(g^\bullet(U);g(U))}{H(g^\bullet(U)) + H(g(U))}
\end{align}
where $U$ is a random variable uniformly distributed on the elements of $\Xcal$. Thus, the \gls{NMI} returns the similarity between the estimated and true partition. It has a lower bound of 0 (independent partitions) and an upper bound of 1 (for identical partitions). To avoid the impact of random initialization on the results, we average \gls{NMI} values over 10 randomized runs.

\begin{figure*}[!thbp]
\centering
\subfigure[][Circles dataset with $\beta=0$]{\includegraphics[width=0.32\textwidth]{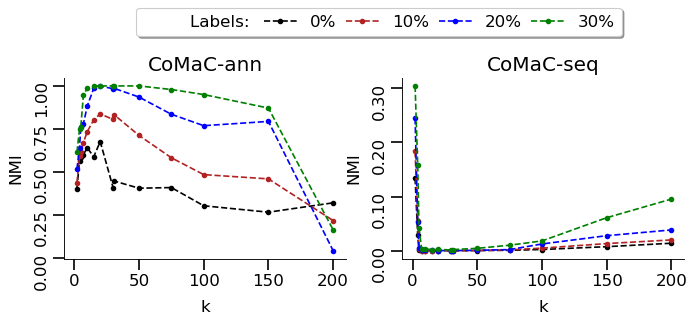}}\hfill
\subfigure[][Circles dataset with $\beta=0.5$]{\includegraphics[width=0.32\textwidth]{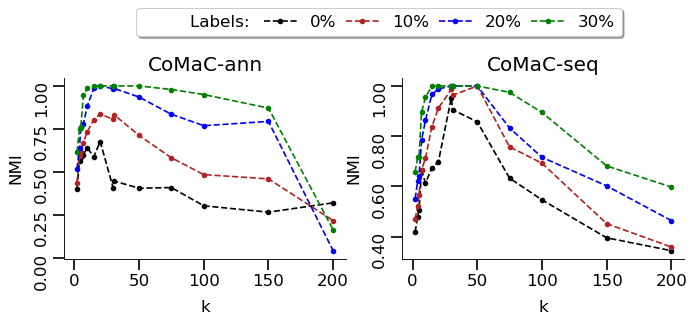}}\hfill
\subfigure[][Circles dataset with $\beta=1$]{\includegraphics[width=0.32\textwidth]{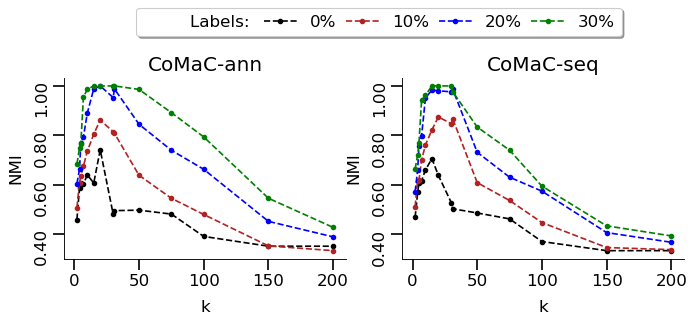}}\\
\subfigure[][Iris dataset with $\beta=0$]{\includegraphics[width=0.32\textwidth,trim={0cm 0cm 0cm 0cm}, clip]{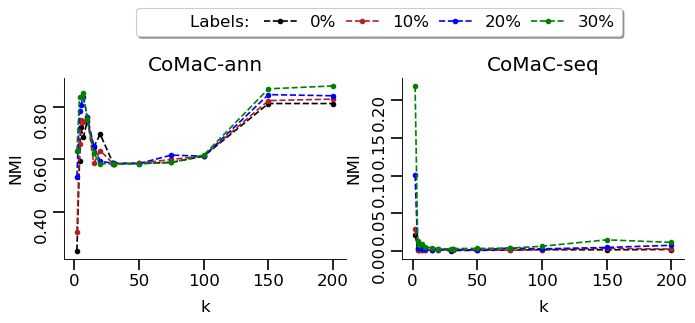}}\hfill
\subfigure[][Iris dataset with $\beta=0.5$]{\includegraphics[width=0.32\textwidth,trim={0cm 0cm 0cm 0cm}, clip]{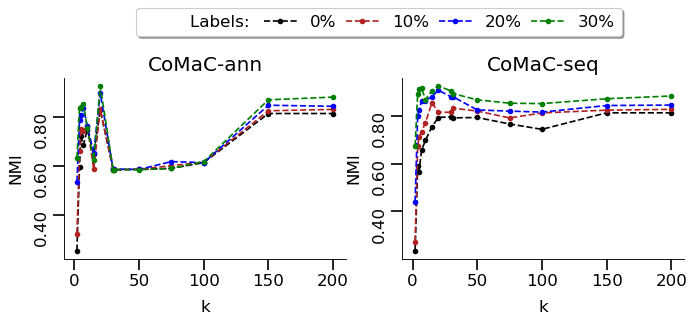}}\hfill
\subfigure[][Iris dataset with $\beta=1$]{\includegraphics[width=0.32\textwidth,trim={0cm 0cm 0cm 0cm}, clip]{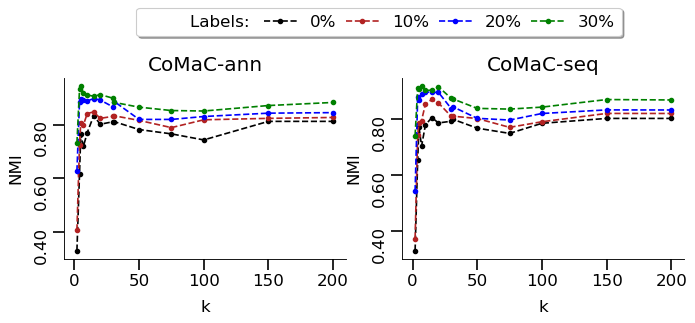}}
\caption{Dependence of the clustering accuracy on the parameter $k$ for the circles dataset (a--c) and the Iris dataset (d--f).}
\label{fig:exp:k_var}
\end{figure*}

\subsection{Sensitivity Analysis of \our{}}
\label{sec:sensitivity}

In this part, we investigate the effect of the hyperparameters selection on the clustering results produced by \gls{MCA}. To be consistent with \cite{smieja_2017}, we generate pairwise constraints from randomly sampled partition-level side information. 

\paragraph{Influence of parameter $k$}

Both for the sequential and annealing algorithm (referred to as \acrshort{MCA-ann} and \acrshort{MCA-seq}) one can observe an influence of the hyperparameter $k$ on the clustering accuracy. Ideally, $k$ is chosen such that the transition probability matrix is nearly completely decomposable, which strongly depends on the chosen dataset. In this subsection, we analyse the influence of $k$ on the three circles dataset shown in Figure~\ref{fig:circles}. Data points are placed uniformly distributed at radii \{0.5, 7, 15\} and corrupted by spherical Gaussian noise with standard deviation of 0.3.

We analyse the performance of \acrshort{MCA-ann} and \acrshort{MCA-seq} with $\beta = \{0,~0.5,~1\}$ for semi-supervised clustering where 0\%, 10\%, 20\%, 30\% of data points are labeled while $k$ is varied (see \autoref{fig:exp:k_var}, top). On the one hand, the experiment shows that \acrshort{MCA-ann} performs more robustly than \acrshort{MCA-seq} w.r.t. the hyperparameter $\beta$. On the other hand, we observe that the additional side information makes \gls{MCA} more robust to the selection of the hyperparameter $k$, at least for this dataset. Clustering accuracy degrades for increasing values of $k$, and the degradation is less severe the more data points are labeled.

The same experiment is repeated for the Iris dataset (see \autoref{fig:exp:k_var}, bottom). As it can be seen, the NMI as a function of $k$ shows less variations than for the three concentric circles. As expected, the optimal value of $k$ depends on the dataset. However, for $k<50$, the performance is quite stable for both datasets and all considered levels of side information. Thus, for all subsequent experiments we set $k=20$ rather than optimize it for each dataset.

\begin{figure}[!thbp]
\centering
\includegraphics[width=0.75\textwidth,trim={0cm 0cm 0cm 1cm}]{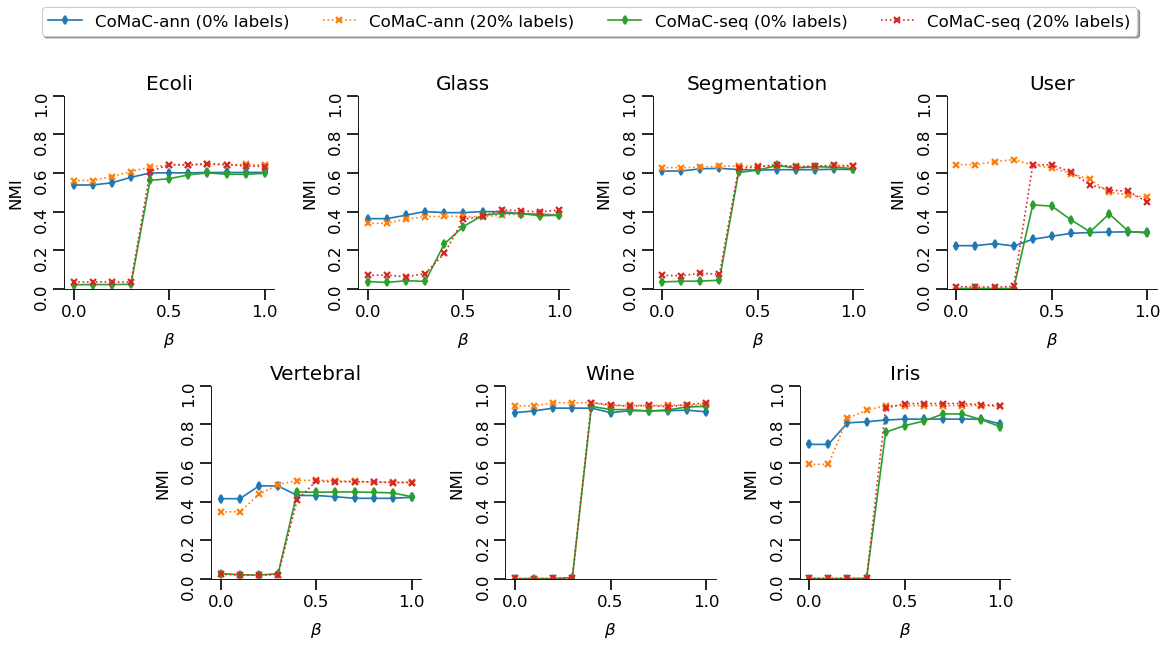}
\caption{Influence of the parameter $\beta$ on the clustering accuracy in the unsupervised and semi-supervised setting (fraction of labeled data = 20\%).}
\label{fig:exp:beta}
\end{figure}

\paragraph{Influence of parameter $\beta$}
We next analyse the influence of the parameter $\beta$ for a constant setting of $k=20$. The results of the experiment for the unsupervised case and for a semi-supervised case where 20\% of the data points are labeled and used to generate the pairwise constraints are shown in Figure~\ref{fig:exp:beta}.

The sequential algorithm performs particularly badly for small $\beta$ values as it is prone to getting stuck in bad local minima. This parallels the behavior of the Markov aggregation method proposed in~\cite{Amjad_2020}. Since these bad minima for small values of $\beta$ cannot be escaped by introducing additional side-information, the annealing scheme was proposed, reducing $\beta$ iteratively.
The annealing algorithm was introduced to ensure a convergence to a good local minimum close to the global minimum. When varying the parameter $\beta$ in smaller step sizes, we can observe that the annealing algorithm returns stable results for the accuracy (see Figure~\ref{fig:exp:beta}). 
For most datasets the additional side-information adds more stability and increased accuracy to the annealing algorithm with respect to variations of $\beta$.
For all following experiments, we choose $\beta = 0.5$, which results in the same cost function as in~\cite{ALUSH2016284}.

\subsection{Evaluation}
\label{sec:evaluation}

\begin{table}[t]
\centering
\caption{Description of the datasets used in our experiments. For datasets with $+$, principal components analysis was used to reduce dimensionality. For Wine dataset we normalize the attributes for \our{}.}
\label{tab:datasets}
\begin{tabular}{lccc}
\hline
\textbf{dataset} & \textbf{\# Instances} & \textbf{\# Features} & \textbf{\# Classes} \\ \hline
Ecoli$^+$         & 327                   & 5                    & 5                   \\
Glass             & 214                   & 9                    & 6                   \\
Iris              & 150                   & 4                    & 3                   \\
Segmentation$^+$  & 210                   & 5                    & 7                   \\
User Modeling     & 403                   & 5                    & 4                   \\
Vertebral         & 310                   & 6                    & 3                   \\
Wine*              & 178                   & 13                   & 3                   \\
\hline
\end{tabular}
\end{table}

Next, we compare \gls{MCA} with the state-of-the-art semi-supervised clustering techniques on several UCI datasets \cite{UCI} as described in Table \ref{tab:datasets}. For a fair comparison, we follow the experimental setup in \cite{smieja_2017} and directly use the clustering results of comparative methods reported there. We assume that the number of clusters is known.

\paragraph{Experimental setup} We consider \acrshort{MCA-ann} and \acrshort{MCA-seq} with $k=20$ and $\beta=0.5$ throughout all experiments. The latter parameter setting corresponds to the clustering method proposed in~\cite{ALUSH2016284}, albeit for a different transition probability matrix $\Pvec$. 

The following baselines are considered (see~\cite{smieja_2017} for a description of hyperparameters selection):
\begin{itemize}
    \item \gls{CEC-IB}~\cite{smieja_2017}: model-based clustering based on cross-entropy and information bottleneck using partition-level side information. We considered two values of a hyperparameter, denoted as CEC-IB$_1$ and CEC-IB$_0$. This is the only method that is initialized with twice the correct number of clusters since it identifies the optimal number automatically. 
    \item mixmod: \gls{GMM} with a partition-level side information implemented in R package Rmixmod~\cite{r_mixmod}.
    \item c-GMM~\cite{cGMM}: \gls{GMM} that uses pairwise constraints. 
    \item k-means~\cite{k_means}: the extension of k-means that supports partition-level side information. 
    \item fc-means~\cite{fuzzy_1,fuzzy_2}: a fuzzy c-means using partition-level side information.
    \item spec~\cite{spec}: a spectral clustering algorithm that incorporates pairwise constraints
\end{itemize}

\begin{figure*}[!thbp]
\centering
\subfigure[The side information covers the elements of all classes\label{fig:exp:all_classes}]{\includegraphics[width=0.75\textwidth]{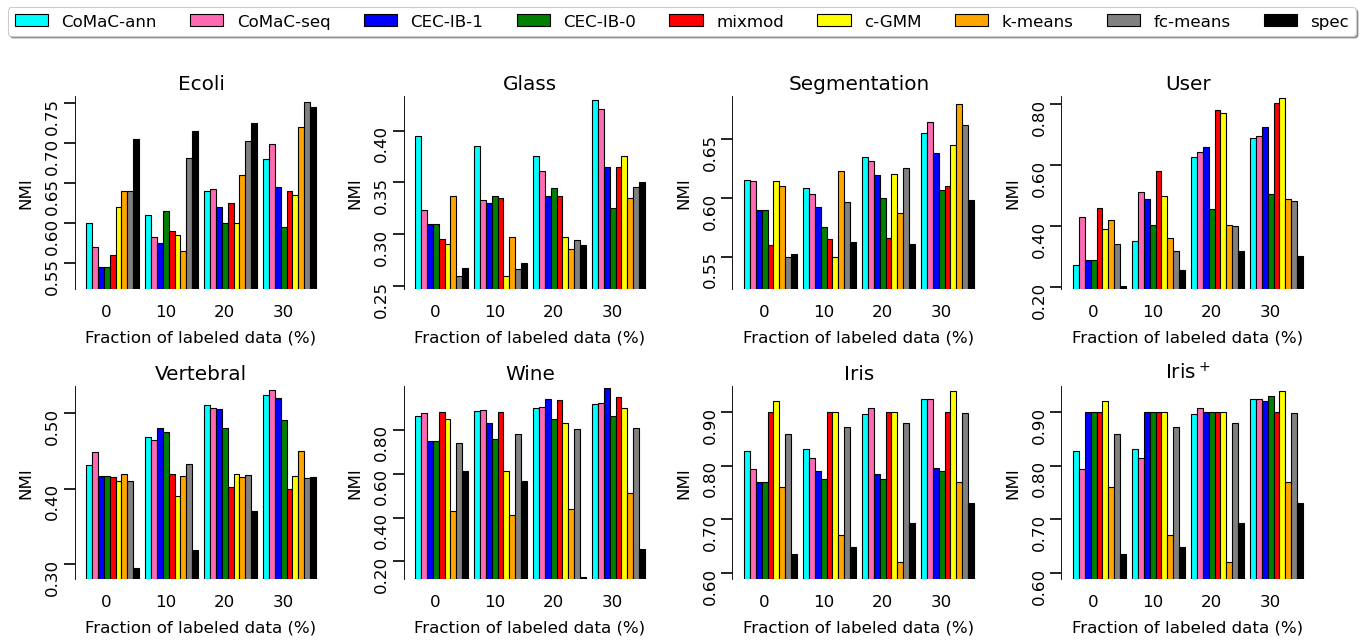}}\\
\subfigure[The side information covers the elements of only two classes\label{fig:exp:2_classes}]{\includegraphics[width=0.75\textwidth,trim={0cm 0cm 0cm 2cm}, clip]{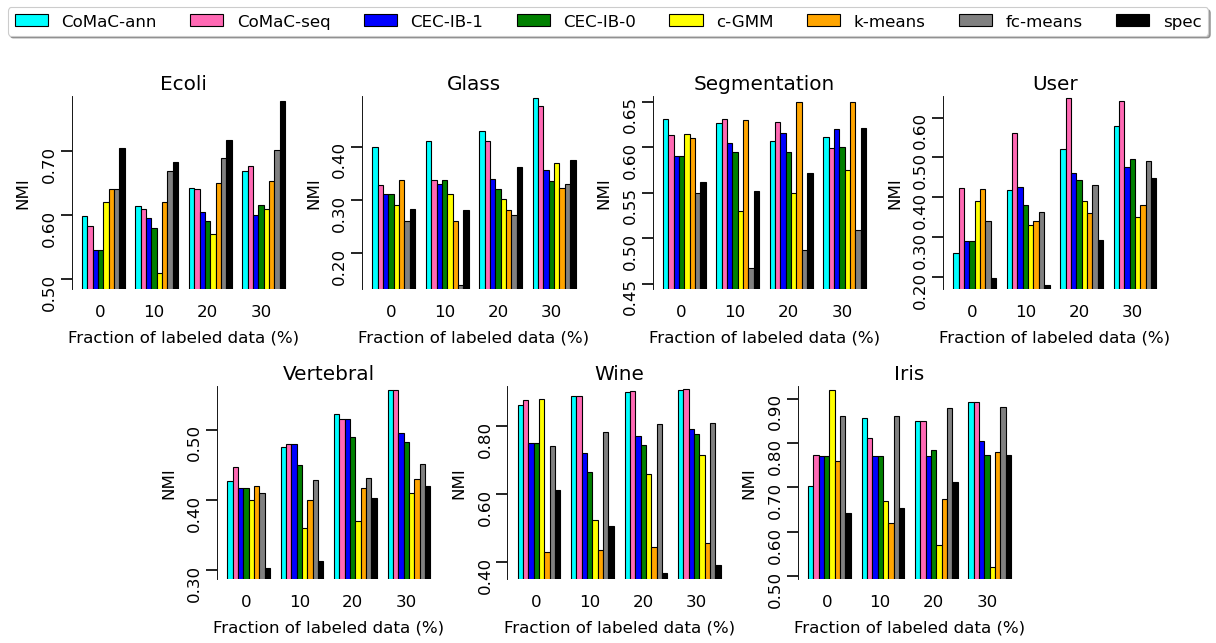}}
\caption{Normalized mutual information computed on UCI datasets with noise-free side information from all classes (a) and two classes only (b). The parameters of \acrshort{MCA} are set to $\beta=0.5$ and $k=20$.
}
\label{fig:exp:comparison}
\end{figure*}

Since some of the comparison methods reported in \cite{smieja_2017} only accept partition-level side information, those methods are provided with the ground truth clusters $g^\bullet(x)$ for a subset $\Xcal'$ of data points. More precisely, the partition-level side information is generated by choosing $0\%, 10\%, 20\%$, and $30\%$ of the data points and labeling them according to their class. This partition-level side information  is subsequently converted to pairwise constraints and incorporated to the remaining methods. To allow for a fair comparison, the constraint sets $\Mcal$ and $\Ncal$ are exhaustive, i.e., they contain all pairwise constraints that are implied by partition-level side information. Specifically, if $|\Xcal'|=m$, then $|\Mcal|+|\Ncal|=m(m-1)/2$.

\paragraph{Clustering with side information from all classes}

First, we consider a typical case, where the partition-level side information covers elements of all classes. Figure~\ref{fig:exp:all_classes} shows the accuracy of the clustering results for each algorithm and dataset for different fractions of labeled data points. Overall, we can observe that \acrshort{MCA} clearly benefits from labeled data, at least for $k=20$ and $\beta=0.5$, as NMI increases with increasing amounts of labeled data points. 

The improvement of \acrshort{MCA} performance due to side information in comparison to the other techniques is most notable on the Iris dataset. Only 20\% of labeled data points noticeably improve the accuracy of \acrshort{MCA} while the other techniques do not benefit as much from additional side information. \acrshort{MCA} furthermore achieves superior performance on the Glass, Segmentation, Vertebral and Wine datasets. Both k-means and spec are sensitive to the scale of attributes, which may partially explain why these methods perform worse on the Wine dataset (\gls{MCA} was run on the normalized Wine dataset). Interestingly, on the Vertebral dataset, all algorithms perform equally well in the unsupervised case. However, when incorporating labeled data, both \acrshort{MCA} and \gls{CEC-IB} outperform all other methods. 

\begin{figure*}[!thbp]
\centering
\includegraphics[width=0.7\textwidth]{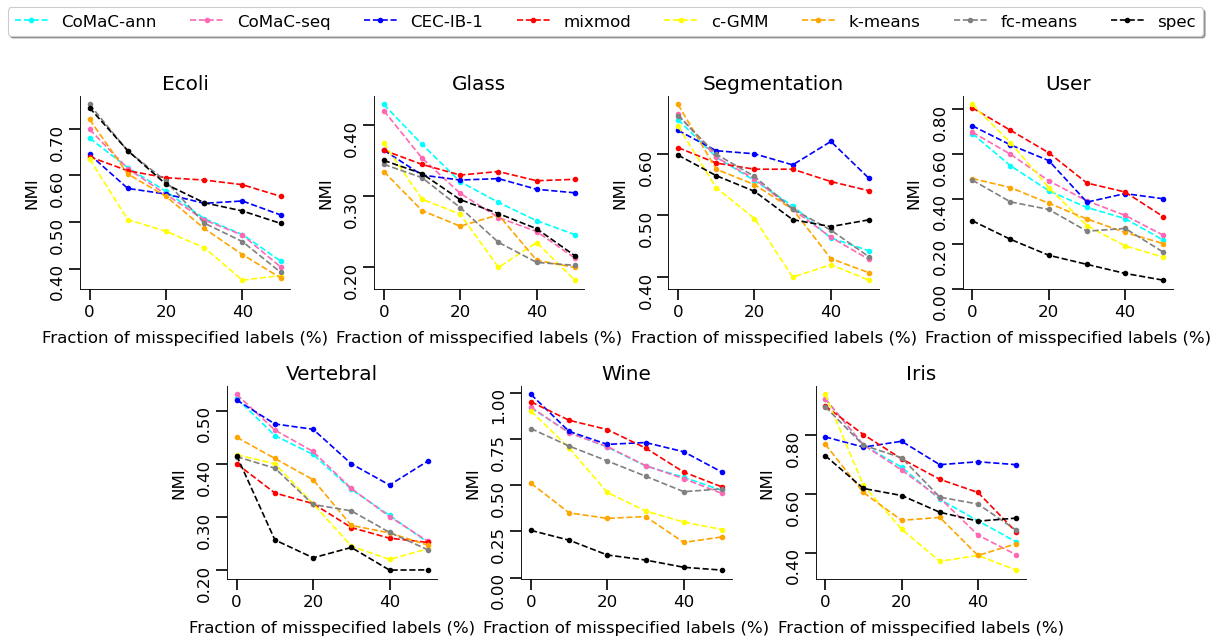}
\caption{Normalized mutual information computed on UCI datasets with noisy partition-level side information from all classes. The parameters of \acrshort{MCA} are set to $\beta=0.5$ and $k=20$.}
\label{fig:exp:noise}
\end{figure*}

\paragraph{Clustering with side information from a subset of classes}

Next, we investigate the case where the side information does not cover all classes. Now, a certain percentage (0\%, 10\%, 20\%, 30\%) of data points from only two classes is selected and used for labeling. Following the work of \cite{smieja_2017}, the two classes must cover at least 30\% of the total data. The goal is to determine the ability of our clustering algorithm to correctly identify all classes, although it is given information about only two of them. 

The results reported in Figure~\ref{fig:exp:2_classes} show that \acrshort{MCA} is robust against missing labels from other classes. The advantage of \gls{MCA} is especially evident in the case of Vertebral, Wine and User dataset, but it also performs well on Glass and Segmentation datasets. Interestingly, \acrshort{MCA-seq} returns significantly better scores than \acrshort{MCA-ann} on the User data.

\subsection{Limitations}\label{sec:limitations}

In this last part, we investigate current limitations of \acrshort{MCA}. Specifically, we examine the effects of erroneous side information and of how pairwise constraints are propagated before being utilized in Algorithm~\ref{alg:myalgo-seq}. Furthermore, we discuss an inherent shortcoming of the greedy coloring initialization in Algorithm~\ref{alg:coloring} concerning cannot-link constraints. The insights of this section thus suggest interesting avenues for future research.

\paragraph{Erroneous partition-level side information} We first examine the robustness of our CoMaC algorithm against erroneous partition-level side information. In this case, 30\% of the data points are labeled, for which a certain percentage (0\%, 10\%, 20\%, 30\%, 40\%, 50\%) of labels are substituted with random incorrect labels.
Note that the pairwise constraints generated from the noisy partition-level side information are not contradictory, but only are inconsistent with the ground-truth. 

When compared to other clustering techniques (see \autoref{fig:exp:noise}), the performance of \acrshort{MCA} is on par with c-GMM, k-means, fc-means and spec, which are all highly sensitive to noise: For four datasets, the semi-supervised setting with only 10\% of erroneous labels is even outperformed by the unsupervised \acrshort{MCA} algorithm. Besides \acrshort{MCA}, c-GMM and spec incorporate side information using pairwise constraints, and both exhibit a steeply falling accuracy as a function of the level of noisy side information.

If not generated from partition-level side information, erroneous pairwise constraints can be conflicting. Consolidating these conflicts either using separate algorithms or via specifically designed cost functions may lead to a degradation in accuracy that is less substantial than for erroneous partition-level side information. Verifying this conjecture will require further experiments and is out of the scope of the current work.

\paragraph{Propagation of pairwise constraints}
As discussed in Section~\ref{sec:method}, the pairwise constraint sets $\Mcal$ and $\Ncal$ may not be exhaustive. When dealing with noise-free pairwise constraints, we may encounter the following two cases: if $(x,x')\in\Mcal$ and $(x,x'')\in\Ncal$, then also $x'$ and $x''$ should not link; and if  $(x,x')\in\Mcal$ and $(x,x'')\in\Mcal$, then also $x'$ and $x''$ should link. The first case is accounted for in the function \textsc{funcCannot}($\Mcal, \Ncal, x$). To study the influence of the propagation of must-link constraints, we compare the propagation function \textsc{funcMust}($\Mcal, \Ncal, x$) as described in Section~\ref{sec:method} with a more primitive version that only returns as $\Mcal_x$ all $x'$ that are linked with $x$ explicitly in $\Mcal$.

\begin{figure}[t]
    \centering
    \subfigure[With cannot-link constraints\label{fig:exp:propagate_Mlinks}]{\includegraphics[width=.3\textwidth,trim={16.25cm 0 0 0},clip]{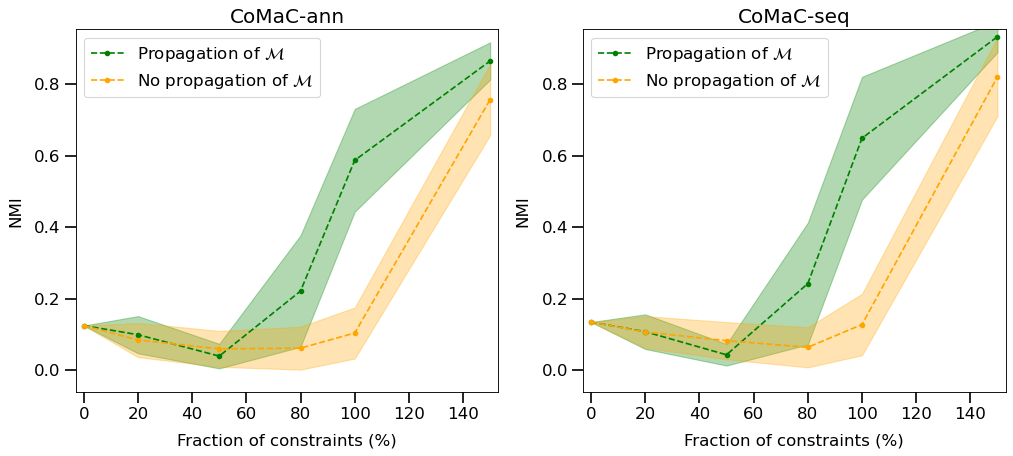}}
    \subfigure[Without cannot-link constraints\label{fig:exp:propagate_Mlinks_no_N}]{\includegraphics[width=.3\textwidth,trim={16.25cm 0 0 0},clip]{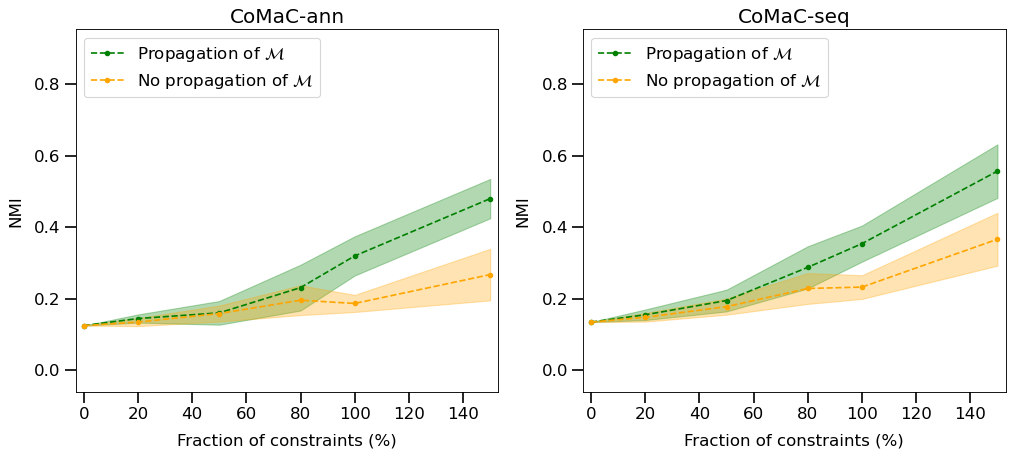}}
    \caption{Influence of propagation of must-link constraints on the accuracy for the Ionosphere dataset (351 samples, 34 features, 2 classes)}
    \label{fig:propagate}
\end{figure}

We performed this experiment on four UCI datasets (Ionosphere, Iris, User, Wine) for a given fraction of pairwise constraints $(|\Mcal|+|\Ncal|)/|X| = \{0\%,20\%,50\%,100\%,150\%\}$. The influence on accuracy due to the propagation of must-link constraints depends on the datasets and could be observed most prominently on the Ionosphere dataset (see Figure~\ref{fig:exp:propagate_Mlinks}). For a small number of pairwise constraints, propagation has no noticeable effect as the randomly sampled must-link constraints rarely overlap. Only after a certain fraction of constraints is added, accuracy starts to increase while simultaneously the influence of propagation on accuracy is visible. However, these effects are highly dependent on the individual datasets. For Iris, User, and Wine, propagation of must-link constraints did not increase accuracy significantly.

\paragraph{Negative influence of cannot-link constraints}
Interestingly, in Figure~\ref{fig:exp:propagate_Mlinks} we could observe an initial drop in accuracy after including 50\% of constraints. There, the accuracy of the clustering on the Ionosphere dataset degrades towards zero, indicating random partitioning. When initializing the graph via a greedy coloring algorithm, vertices connected by cannot-link constraints are assigned to the first cluster available. As the Ionosphere dataset consist of two classes only, those constrained data pairs are forced to stay in place during the rest of the algorithm as there are no further free classes available. Excluding all cannot-link constraints avoids the initial drop in accuracy and leads to a monotonically increasing accuracy (see Figure~\ref{fig:exp:propagate_Mlinks_no_N}). Additionally, we could observe that the number of constraints where the propagation of must-link constraints starts to improve clustering accuracy seems to corresponds to the point where cannot-link constraints stop harming the performance. However, at least for Ionosphere and a larger amount of constraints, the performance without cannot-link constraints was generally lower than by considering them.

Again, these effects depended on the individual datasets. To gain further into the reason behind these effects and possible mitigation by adapting the algorithms, further investigation is required. 

\section{Discussion and Conclusion}

In this work we have extended the unsupervised optimization algorithm for clustering via Markov aggregation of \cite{Amjad_2020} to accept pairwise constraints. We showed that that the use of pairwise constraints successfully lowers the algorithm's sensitivity to hyperparameter settings. Extensive experiments using pairwise constraints from partition-level side information confirmed that our algorithm can learn non-linear decision boundaries between clusters and competes with state-of-the-art semi-supervised clustering techniques. Especially when provided with side information covering only a subset of all classes, our method can achieve better results than existing techniques.

Finally, a number of potential limitations require further work beyond the scope of this paper. First, our method reacts sensitively to noisy constraints generated from erroneous partition-level side information. Second, presented non-exhaustive pairwise constraint sets, propagating the constraints over the whole dataset can improve performance. However, this task becomes non-trivial when pairwise constraints are inconsistent due to noise. Finally, when dealing with sparse pairwise constraints, we could observe that the greedy graph initialization hampers clustering performance for low to moderate numbers of cannot-link constraints. Future work shall determine possible explanations for these limiting factors and subsequently adapt the algorithm to mitigate said problems. 

\begin{acks}
  The work of Sophie Steger has been supported by iDev40. The iDev40 project has received funding from the ECSEL Joint Undertaking (JU) under grant agreement No 783163. The JU receives support from the European Union’s Horizon 2020 research and innovation programme. It is co-funded by the consortium members, grants from Austria, Germany, Belgium, Italy, Spain and Romania. 
  
  The work of Bernhard C. Geiger has been supported by the HiDALGO project and has been funded by the European Commission's ICT activity of the H2020 Programme under grant agreement number 824115. 
  
  The Know-Center is funded within the Austrian COMET Program - Competence Centers for Excellent Technologies - under the auspices of the Austrian Federal Ministry of Climate Action, Environment, Energy, Mobility, Innovation and Technology, the Austrian Federal Ministry of Digital and Economic Affairs, and by the State of Styria. COMET is managed by the Austrian Research Promotion Agency FFG.
\end{acks}

\bibliographystyle{ACM-Reference-Format}
\bibliography{additional_references} 

\end{document}